\documentclass[11pt,a4paper]{article}
\usepackage{times,latexsym}
\usepackage{url}
\usepackage[T1]{fontenc}
\usepackage{graphicx}
\usepackage{booktabs}
\usepackage{CJKutf8}
\usepackage{xspace}

\newcommand{\arukikata}{Arukikata.~Co.,~Ltd.\xspace}

 \usepackage[acceptedWithA]{tacl2021v1}
\usepackage{xspace,mfirstuc,tabulary}

\newif\iftaclinstructions
\taclinstructionsfalse % AUTHORS: do NOT set this to true
\iftaclinstructions
\renewcommand{\confidential}{}
\renewcommand{\anonsubtext}{(No author info supplied here, for consistency with
TACL-submission anonymization requirements)}
\newcommand{\instr}
\fi

\iftaclpubformat % this "if" is set by the choice of options

\else

\fi

\title{NAIST Academic Travelogue Dataset}

\author{Hiroki Ouchi$^{\clubsuit,\heartsuit,}$\Thanks{Corresponding author.} \hspace{0.4cm} Hiroyuki Shindo$^{\clubsuit}$  \hspace{0.4cm} Shoko Wakamiya$^{\clubsuit}$  \hspace{0.4cm} \textbf{Yuki Matsuda}$^{\clubsuit,\heartsuit}$\\   \hspace{0.4cm} \textbf{Naoya Inoue}$^{\spadesuit}$ \hspace{0.4cm} \textbf{Shohei Higashiyama}$^{\diamondsuit,\clubsuit}$ \hspace{0.4cm} \textbf{Satoshi Nakamura}$^{\clubsuit}$ \hspace{0.4cm} \textbf{Taro Watanabe}$^{\clubsuit}$\\
  {$^\clubsuit$ NAIST} \hspace{0.3cm} {$^\spadesuit$ JAIST} \hspace{0.3cm} {$^\diamondsuit$ NICT} \hspace{0.3cm} {$^\heartsuit$ RIKEN}\\
  \{\texttt{hiroki.ouchi,shindo,wakamiya,yukimat\}@is.naist.jp}, \\ \texttt{naoya-i@jaist.ac.jp}, \hspace{0.25cm}\texttt{shohei.higashiyama@nict.go.jp}, \\ \{\texttt{s-nakamura,taro\}@is.naist.jp}}

\date{}

\begin{document}
\maketitle
\begin{abstract}
We have constructed NAIST Academic Travelogue Dataset (ATD) and released it free of charge for academic research.
% This dataset contains Japanese text data, including 4,500 domestic travelogues and 9,500 overseas travelogues, with a total of over 31 million words.
This dataset is a Japanese text dataset with a total of over 31 million words, comprising 4,672 Japanese domestic travelogues and 9,607 overseas travelogues.
Before providing our dataset, there was a scarcity of widely available travelogue data for research purposes, and each researcher had to prepare their own data.
%Because very few travelogue datasets have been available for research purposes so far, each researcher had to prepare his or her own data.
% This makes it difficult to replicate existing studies and to conduct fair comparative analysis of experimental results.
This hinders the replication of existing studies and fair comparative analysis of experimental results.
Our dataset enables any researchers to conduct investigation on the same data and to ensure transparency and reproducibility in research.
%By making our travelogue dataset open for research, we have enabled researchers to use the same data, thus ensuring transparency and reproducibility in research.
% researchers can use the same data, which ensures transparency and reproducibility in research.
In this paper, we describe the academic significance, characteristics, and prospects of our dataset.
\end{abstract}

\section{Introduction}
\label{sec:introduction}
The COVID-19 outbreak has drawn more attention to the dynamics between humans and places.
In particular, information regarding the level of a crowd (human concentration) in certain places and the movement between places (human mobility) is crucial for decision-making on promoting or restricting activities, regardless of the scale of societies, including government, local communities, or individuals.
%In particular, location information of traveling activities is highly valuable for government decision-making, regional sightseeing promotion and personal travel planning.
Under these circumstances, we have explored methodologies for analyzing human behavior from the perspective of \textit{places} and adopted ``text'' as a valuable resource for such analysis.
%In the Corona Disaster, information on the degree of congestion at a given location (human concentration) and the traffic between locations (human movement) is extremely important for decision making to promote/control behavior, regardless of the level of granularity, such as government, local government, or individuals.
%In this paper, we explore methodologies for analyzing human behavior in terms of location.
%As a resource for the analysis, we have adopted ``text.'' 
%We aim to develop a computer that can recognize the places where persons appearing in text behave, and ground the places on a map of the real world.
Specifically, our objective is to develop a computer system that can accurately recognize the places where characters engage in activities, and ground these places onto a real-world map.
As a first step toward this goal, we have constructed NAIST Academic Travelogue Dataset (ATD).
The original travelogue documents were collected from an online travelogue posting service, and the dataset with the auto-analyzed information is distributed free of charge to academic research institutions for research purposes.\footnote{The NAIST Academic Travelogue Dataset (ATD) was previously provided as the Arukikata Travelogue Dataset (\url{http://doi.org/10.32130/idr.18.1}) through the NII IDR and has now been redistributed by NAIST.}

\paragraph {Why use ``text?''}
We can track human locations by utilizing the Global Positioning System (GPS) functionality in mobile devices like smartphones.
% If you only want to grasp where people visited, you can do so by 
%We can obtain information on points that people visited using the Global Positioning System (GPS) functionality of mobile devices.
However, it is challenging to grasp the mutual relationship between humans and places solely from GPS data.
The relationship, for instance, includes human activities in a particular place, the subjective value attributed to that place, and the impressions and sensations evoked by being there.
%from GPS data, it is difficult to understand the dynamics between humans and places, such as human behavior in a place, the value assigned to a place, and the impressions and sensations received from a place.
Such information plays an important role in analyzing the dynamics of human activities and environmental conditions, especially in geography, tourism studies, and cultural anthropology.
Text is a typical resource containing this type of information.
Structuring and organizing texts enable the extraction of such valuable information.
For this reason, we adopted text data as our target.

\begin{figure*}[t]
  \begin{center}
    \includegraphics[width=15.9cm]{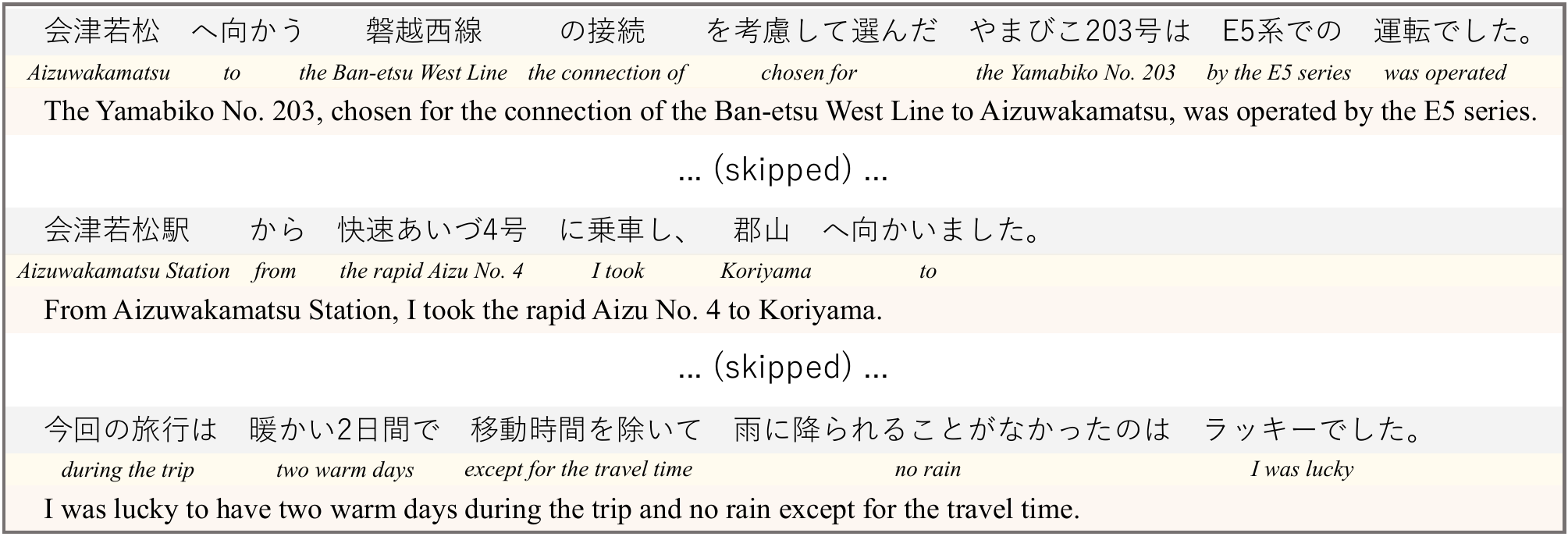}
  \end{center}
  \vspace{-0.5cm}
  \caption{Example of a travelogue.}
  \label{fig:example_of_travelogue}
\end{figure*}

\begin{figure}[t]
  \begin{center}
    \includegraphics[width=7cm]{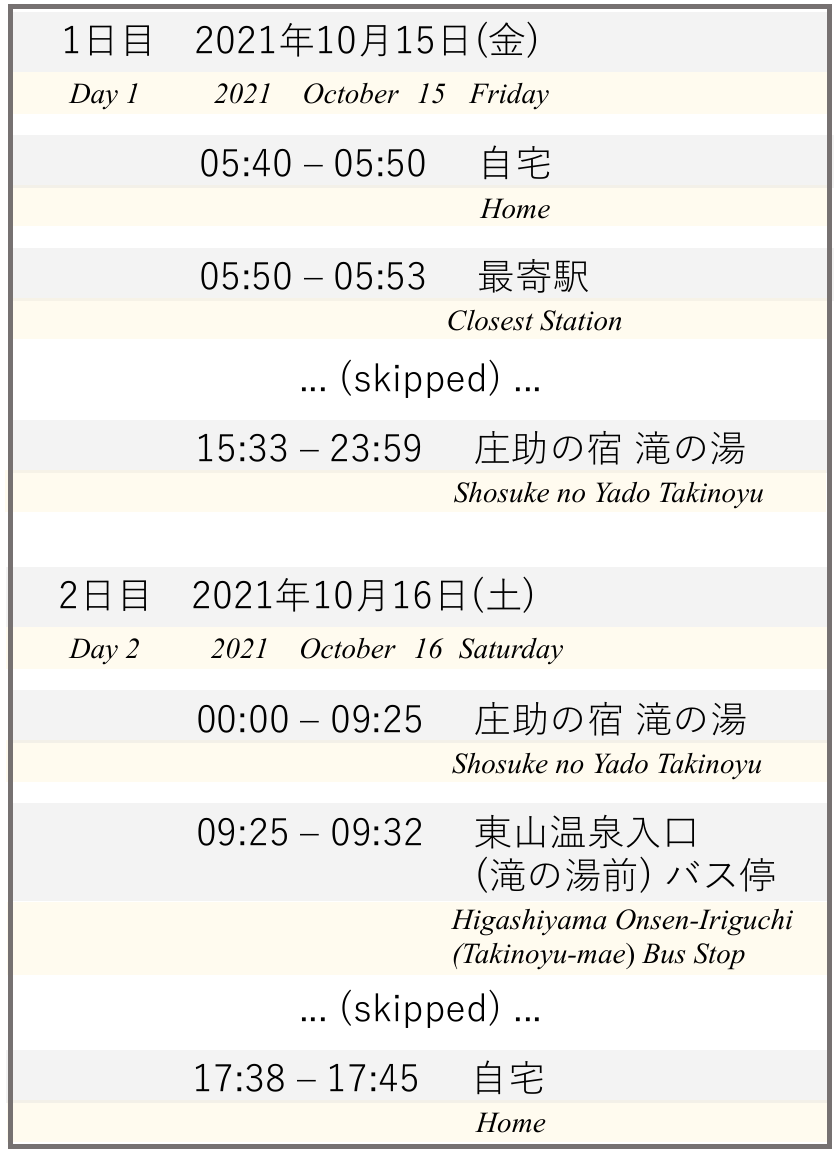}
  \end{center}
  \vspace{-0.4cm}
  \caption{Example of a travel schedule.}
  \label{fig:example_of_schedule}
\end{figure}

\paragraph{Why use ``travelogues?''}
%Existing studies that analyze travelogues often frame their research from the perspective of the relationship between tourists and tourist spots~\cite{hao2009,hao2010,pang2011}.
Existing studies that analyze travelogues often set their research objectives from the perspective of the relationship between tourists and tourist spots~\cite{hao2009,hao2010,pang2011}.
In contrast, we adopt a more abstract perspective; how do \textit{humans} and \textit{places}, the fundamental elements of the real world, interact with each other, and how are these interactions described in text?
A typical genre of text that includes such content is \textit{travelogues}.
Human-place interactions are often described in other genres of text, such as novels, news articles, and SNS posts.
We have chosen ``travelogues'' as a starting point with the prospect to analyze a wider variety of text.

\paragraph{Academic significance of our dataset}
Travelogues have been frequently used as a valuable resource for text mining~\cite{akehurst2009}.
Especially in the field of tourism informatics, travelogues have been used for analyzing the reputation of each place or facility and extracting various types of information.
However, widely available travelogue datasets were scarce for research purposes, and each researcher had to prepare their own data.\footnote{The main reason for this is that travelogues are user-generated contents and usually cannot be redistributed unless various copyright-related requirements are met.}
As a result, replicating and comparing studies has become challenging.
%However, it has been difficult to replicate studies and to conduct fair comparative analysis of experimental results because researchers often use the travelogues obtained from some web cites by themselves\footnote{The main reason for this is that travelogues are user generated contents and usually cannot be redistributed unless various copyright-related requirements are met.}.
To remedy the situation, we have made our travelogue dataset available for research under certain conditions, enabling researchers to use the same data.
Such shared use of datasets ensures transparency and reproducibility in research and facilitates comparisons with other studies~\cite{osuga2021}.
%By making our travelogue dataset open for research under certain conditions, as \citet{osuga2021} point out, researchers can use the same data, which ensures transparency and reproducibility of the research, and facilitating comparisons with other studies.
This use of datasets can also promote open science, accelerate the accumulation of research findings, and foster continued growth in the field of research.
%It will advance open science and accelerate the accumulation of research findings, which in turn will further develop research.

%%%%%%%%%%%%%%%%%%%%%%%%%%%%%%%%%%%%%%%%%%%%%%%%%%%%%%%%%%%%%%%%%%%%%%%%
\section{Dataset}
Our dataset is Japanese text data posted on the travelogue posting service\footnote{\url{https://tabisuke.arukikata.co.jp/} (This service was terminated in March 2022)} on the website operated by \arukikata from November 2007 to February 2022.
Each post comprises a \textbf{travelogue} (Sec.~\ref{sec:travelogue}) and its \textbf{travel schedule} (Sec.~\ref{sec:schedule}).

\begin{table*}[t]
  \centering
  \begin{tabular}{l | cc | cc | c} \toprule
  		  & \multicolumn{2}{c|}{Domestic} & \multicolumn{2}{c|}{Oversea} & All \\ 
		  & Schedule & w/o Schedule  & Schedule & w/o Schedule & \\ \midrule
                 Travelogues & \multicolumn{1}{r}{3,153} & \multicolumn{1}{r|}{1,519} & \multicolumn{1}{r}{6,419}  & \multicolumn{1}{r|}{3,188} & \multicolumn{1}{r}{14,279} \\
                 Paragraphs & \multicolumn{1}{r}{76,307} & \multicolumn{1}{r|}{16,412} & \multicolumn{1}{r}{188,908}  & \multicolumn{1}{r|}{58,700} & \multicolumn{1}{r}{340,327} \\
                 Characters & \multicolumn{1}{r}{5,878,704} & \multicolumn{1}{r|}{1,541,124} & \multicolumn{1}{r}{19,273,201}  &  \multicolumn{1}{r|}{4,870,061} & \multicolumn{1}{r}{31,563,090} \\
                 Words &  \multicolumn{1}{r}{3,568,354} & \multicolumn{1}{r|}{928,936} & \multicolumn{1}{r}{10,950,950}  &  \multicolumn{1}{r|}{2,785,494} & \multicolumn{1}{r}{18,233,734} \\ 
                 NEs & \multicolumn{1}{r}{304,606} & \multicolumn{1}{r|}{71,598} & \multicolumn{1}{r}{928,487}  &  \multicolumn{1}{r|}{227,191} & \multicolumn{1}{r}{1,531,882} \\
                 POIs & \multicolumn{1}{r}{95,282} & \multicolumn{1}{r|}{25,455} & \multicolumn{1}{r}{268,417} & \multicolumn{1}{r|}{71,830} & \multicolumn{1}{r}{460,984} \\ \midrule
                 Paragraphs/Travelogue & \multicolumn{1}{r}{24.2} & \multicolumn{1}{r|}{10.8} & \multicolumn{1}{r}{29.4}  & \multicolumn{1}{r|}{18.4} & \multicolumn{1}{r}{23.8} \\
                 Characters/Travelogue & \multicolumn{1}{r}{1864.4} & \multicolumn{1}{r|}{1014.5} & \multicolumn{1}{r}{3002.5}  &  \multicolumn{1}{r|}{1527.6} & \multicolumn{1}{r}{2210.4} \\
                 Words/Travelogue &  \multicolumn{1}{r}{1131.7} & \multicolumn{1}{r|}{611.5} & \multicolumn{1}{r}{1706.0}  &  \multicolumn{1}{r|}{873.7} & \multicolumn{1}{r}{1276.9} \\ 
                 NEs/Travelogue & \multicolumn{1}{r}{96.6} & \multicolumn{1}{r|}{47.1} & \multicolumn{1}{r}{144.6}  &  \multicolumn{1}{r|}{71.2} & \multicolumn{1}{r}{107.2} \\
                 POIs/Travelogue & \multicolumn{1}{r}{30.2} & \multicolumn{1}{r|}{16.7} & \multicolumn{1}{r}{41.8} & \multicolumn{1}{r|}{22.5} & \multicolumn{1}{r}{32.2} \\
                 \bottomrule
  \end{tabular}
  \caption{Descriptive statistics. ``NEs'' stands for ``named entities,'' and ``POIs'' stands for ``points of interest'' (i.e., places and facilities). ``w/o Schedule'' stands for ''without travel schedule.''}
  \label{tab:data-stats}
\end{table*}

\subsection{Travelogues}
\label{sec:travelogue}
Travelogues are written about experiences in (i) \textit{domestic travel} or (ii) \textit{overseas travel}.
Table~\ref{fig:example_of_travelogue} illustrates an example of a domestic travelogue.
Generally, travelogues are written in the first-person perspective from the viewpoint of the author.
Readers can adopt the author's viewpoint and experience simulated travel.

Each travelogue has a substantial volume, allowing for discourse elements that span sentences and paragraphs, such as sequences of the author's actions, scene transitions, and coreference relations.
%Since each travelogue has a relatively large volume, there appear diverse inter-sentential linguistic phenomena, such as coreference relations, sequences of actions, and scene transitions.
This characteristic distinguishes travelogues from short texts posted on SNS like Twitter.
%This is a characteristic that distinguishes travelogues from posts on social networking services, such as Twitter posts.
As general content, travelogues depict the daily traveling activities of authors and are suitable for the temporal analysis of actions.
They also cover diverse content, such as descriptions of places, landscapes, and personal impressions, which opens up possibilities for various applications.
%Typical travelogues describe travel activities of a day, which implies that travelogues can be a suitable resource for temporal analysis of human activities.
%They also describe a variety of other contents, such as descriptions of places and landscapes, and impressions of each place and the entire itinerary, which may lead to a wide range of applications.

\subsection{Travel Schedules}
\label{sec:schedule}
Table~\ref{fig:example_of_schedule} illustrates an example of a travel schedule.
One of the key elements is the places the author visited.
For example, in Table~\ref{fig:example_of_schedule}, ``\textit{Home},'' ``\textit{Closest Station},'' and ``\textit{Shosuke no Yado Takinoyu}'' correspond to actual places.
Note that the input of the travel schedule is optional.
%Note that each author freely describes the schedules.
This means that the places mentioned in a travelogue are not guaranteed to be listed in the travel schedule.
Conversely, there may be places listed in the travel schedule that are not mentioned in the travelogue.
%This means that the places mentioned in a travelogue are not necessarily mentioned in the schedule, and the places mentioned in a schedule are not necessarily mentioned in the travelogue.

Another key element is the time period in which the author stayed at each place.
For example, in Table~\ref{fig:example_of_schedule}, ``05:40 - 05:50'' and ``05:50 - 05:53'' correspond to time periods.
Note that there exists no particular input format for the time period information, thus allowing for vague descriptions like ``Morning'' or ``Evening.''
%Note that the information about time periods is freely described by the author, allowing for vague (coarse-grained) descriptions like ``Morning'' or ``Evening.''
%Note that each author freely describes the time periods, so that there are also vague (coarse-grained) expressions, such as ``\textit{Morning}'' or ``\textit{Evening}.''
Taking these key elements into account, the dataset is useful for analyzing human daily activities not only from a \textit{place} perspective but also from a \textit{time} perspective.
%Such information enables to analyze traveling activities in terms of time and place.

\subsection{Descriptive Statistics}
\label{sec:data-stats}
Travelogues in our dataset are divided into ``domestic travelogues'' and ``overseas travelogues,'' and further into those ``with travel schedules'' and those ``without travel schedules.''
We performed word segmentation and named entity extraction for all travelogues by using GiNZA\footnote{We used \texttt{ja\_ginza\_electra} version $5.1.2$.
For details, see the following page: \url{https://github.com/megagonlabs/ginza}.}, a Japanese NLP open source library.
Table~\ref{tab:data-stats} shows the descriptive statistics.
Each cell in the row ``Travelogue'' represents the number of travelogues (articles).
In the same way, each cell in the rows ``Paragraphs,'' ``Characters,'' ``Words\footnote{Word counts vary depending on the word definition (word segmentation criteria). For our dataset, we used the GiNZA's word segmentation mode \texttt{C} (long unit words) for word segmentation.},'' ``NEs,'' and ``POIs\footnote{POIs stands for ``points of interest,'' location-related entities including places and facilities. Among the named entities recognized by using GiNZA, we counted \texttt{LOC}, \texttt{GPE}, or \texttt{FAC} as POIs.}'' indicates each number, respectively.
``*/Travelogue'' is the average number per travelogue.
For example, ``Paragraphs/Travelogue = 24.2'' is that the average number of paragraphs per travelogue is 24.2.

One of the notable characteristics of this dataset is that the average number of characters per travelogue (as indicated in the row ``Characters/Travelogue'') is about 2,000, indicating a substantial amount of text in contrast to shorter posts commonly found on SNS like Twitter.
Another characteristic is that each travelogue contains many POIs, i.e., about 30 place or facility names, as indicated in the row ``POIs/Travelogue.''
This indicates that the dataset includes a large number of place-related expressions and possesses desirable properties for conducting analyses related to places.
%This implies that this dataset is useful for analysis of places.

%%%%%%%%%%%%%%%%%%%%%%%%%%%%%%%%%%%%%%%%%%%%%
\subsection{Prefectures Covered in Domestic Travelogues}
\begin{figure}[t]
  \begin{center}
    \includegraphics[width=8.5cm]{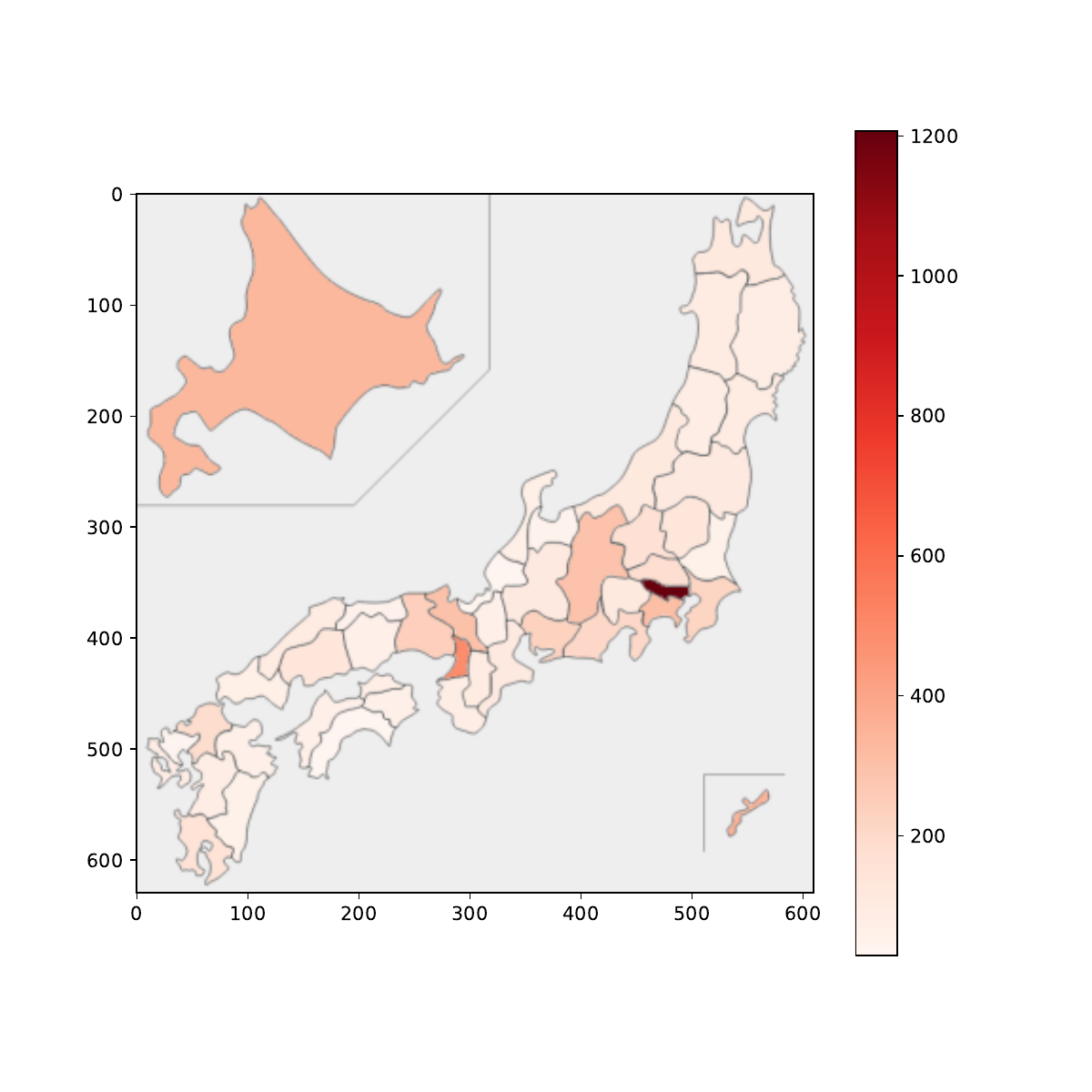}
  \end{center}
  \vspace{-1cm}
  \caption{Distribution of the number of domestic travelogues mentioning each prefecture.}
  \label{fig:japan}
\end{figure}

\begin{table}[t]
  \centering
  \small{
  \begin{tabular}{l | rr } \toprule
  		Prefectures & Travelogues & Percentage \\ \midrule
Tokyo & 1,206 &  25.81\% \\
Osaka & 492 &  10.53\% \\
Okinawa & 353 &   7.56\% \\
Hokkaido & 336 &   7.19\% \\
Kanagawa & 313 &   6.70\% \\
Kyoto & 299 &   6.40\% \\
Nagano & 295 &   6.31\% \\
Hyogo & 242 &   5.18\% \\
Aichi & 231 &   4.94\% \\
Chiba & 219 &   4.69\% \\
    \bottomrule
  \end{tabular}
  }
  \caption{Prefectures ranking. The ``Travelogues'' column indicates the number of travelogues mentioning each prefecture. The ``Percentage'' column indicates the proportion of those travelogues out of the total number of domestic travelogues (4,762).}
  \label{tab:top10-domestic}
\end{table}

Figure~\ref{fig:japan} shows the distribution of domestic travelogues referring to each prefecture.\footnote{Note that the total number is larger than the total number of domestic travelogues (4,672) because some travelogues refer to more than one prefecture.}
One notable characteristic is that all prefectures are covered, which enables analysis of trends in travel activities for each prefecture.
Another characteristic is that Tokyo is most frequently mentioned in travelogues.
In more detail, Table~\ref{tab:top10-domestic} shows the ranking of the top 10 prefectures.
Tokyo is mentioned in 1,206 travelogues, which is more than double the number for Osaka, the second most mentioned prefecture.
However, this does not necessarily imply that Tokyo is the most popular travel destination.
Many travelogues cover the entire travel process, including the starting point, transits, destinations, and ending point.
Tokyo is often mentioned as either the starting point or a transit point in these travelogues, resulting in Tokyo being mentioned most frequently.
Fukui is ranked lowest in the ranking and is mentioned in 29 travelogues.

%%%%%%%%%%%%%%%%%%%%%%%%%%%%%%%%%%%%%%%%%%%%%
\subsection{Countries and Regions Covered in Overseas Travelogues}
\begin{table}[t]
  \centering
  \small{
  \begin{tabular}{l | rr } \toprule
  		Country \& Region & Travelogues & Percentage \\ \midrule
United States & 732 &   7.62\% \\
South Korea & 708 &   7.37\% \\
France & 697 &   7.26\% \\
China & 692 &   7.20\% \\
Taiwan & 611 &   6.36\% \\
Germany & 577 &   6.01\% \\
Thailand & 554 &   5.77\% \\
Italy & 548 &   5.70\% \\
Spain & 413 &   4.30\% \\
Switzerland & 411 &   4.28\% \\
    \bottomrule
  \end{tabular}
  }
  \caption{Countries and regions ranking. The ``Travelogues'' column indicates the number of travelogues mentioning each country or region. The ``Percentage'' column indicates the proportion of those travelogues out of the total number of overseas travelogues (9,607).}
  \label{tab:top10-oversea}
\end{table}

The overseas travelogues in our dataset cover more than 150 countries and regions worldwide, enabling analysis of travel trends in different countries and regions.
Table~\ref{tab:top10-oversea} presents a ranking of the number of travelogues mentioning each country or region.
Unlike Tokyo in Table~\ref{tab:top10-domestic}, there are no particular countries or regions with outstandingly higher numbers of travelogues.
%there is no country or region with an outstandingly large number of travelogues.
In addition, the top-ranked countries and regions align closely with ``Japanese Overseas Travelers by Destination (Visitor Arrivals from Japan)\footnote{\url{https://www.jnto.go.jp/statistics/data/pdf/20220610_4.pdf}}'' provided by Japan National Tourism Organization (JNTO).

%%%%%%%%%%%%%%%%%%%%%%%%%%%%%%%%%%%%%%%%%%%%%%%%%%%%%%%%%%%%%%%%%%%%%%%%
\section{Related Work}
\begin{table}[t]
  \centering
  \small{
  \begin{tabular}{l | crr } \toprule
  		  & Lang & Articles & Words \\ \midrule
		  \scriptsize{Diachronic News and Travel} & En & 23 & 30,747\\
		  \scriptsize{The SpaceBank Corpus} & En &  44 & 21,048 \\
		  \scriptsize{KNB Corpus} & Ja &  91 & 24,900 \\
                 \bottomrule
  \end{tabular}
  }
  \caption{Existing travelogue datasets.}
  \label{tab:data-comp}
\end{table}

There are very few contemporary travelogue datasets available for academic research purposes.
Table~\ref{tab:data-comp} presents a few of these rare exceptions.
The Diachronic News and Travel Corpus\footnote{\url{https://github.com/tommasoc80/DNT}.}\cite{caselli_sprugnoli_dnt2021} includes English texts from three domains (News, Travel Reports, and Travel Guides) divided into two time periods (1862-1939 and 1998-2017).
Table~\ref{tab:data-comp} specifically provides information on the ``Travel Reports'' texts from the ``Contemporary (1998-2017)'' period.
The SpaceBank Corpus\footnote{\url{https://alt.qcri.org/semeval2015/task8/index.php?id=data-and-tools}.} ~\cite{pustejovsky2013capturing} is an annotated corpus including spatial information and was used in SemEval-2015 Task 8~\cite{pustejovsky-etal-2015-semeval}.
Table~\ref{tab:data-comp} provides information on a subset constructed from the travel blog ``Ride for Climate'' entries.
The KNB Corpus\footnote{\url{https://nlp.ist.i.kyoto-u.ac.jp/kuntt/}.}~\cite{hashimoto2011knbc} comprises Japanese texts from four domains (``Kyoto Tourism,'' ``Mobile Phones,'' ``Sports,'' and ``Gourmet'').
The corpus is annotated with linguistic information such as morphological analysis, dependency parsing, predicate-argument structures, ellipsis, and coreference relations.
Table~\ref{tab:data-comp} provides information on the texts from ``Kyoto Tourism.''
These datasets, including manual annotations, cannot be compared to our dataset in terms of ``quantity.''
In the future, we plan to apply diverse annotations to our dataset and further expand its applicability.

%%%%%%%%%%%%%%%%%%%%%%%%%%%%%%%%%%%%%%%%%%%%%%%%%%%%%%%%%%%%%%%%%%%%%%%%
\section{Conclusion}
We have built and released NAIST Academic Travelogue Dataset (ATD).
In this paper, we have primarily discussed the academic significance and characteristics of the dataset.
In this section, we will outline our prospects.
We plan to annotate linguistic information to the dataset.
Specifically, (1) information regarding language expressions about places and (2) information concerning the interaction between humans and places are highlighted.
For (1), we will comprehensively cover language expressions about places, including not only proper
noun phrases such as place names and facility names, but also general noun phrases such as ``this shop'' and ``the restaurants.''
Furthermore, intending to connect with the real world, we are also considering linking each expression in the dataset to map coordinates and geographic databases.
For (2), we will provide information about human ``actions,'' ``thoughts,'' and ``emotions'' in specific places.
We aim to develop tools to extract such information and place information and apply them to various applications. 
%To automatically extract such information, we will develop tools and make use of them for various applications.
Possible applications include analyzing travelers' movement patterns, analyzing trends of tourist destinations, discovering hidden tourist spots, and utilizing the information for travel planning and recommendation.
As mentioned above, various possibilities can be envisioned.
We hope that many researchers in diverse fields will make use of our dataset and advance innovative research and development.

%%%%%%%%%%%%%%%%%%%%%%%%%%%%%%%%%%%%%%%%%%%%%%%%%%%%%%%%%%%%%%%%%%%%%%%%%%%
\section*{Acknowledgments}
We would like to express our deep gratitude for the invaluable cooperation of Yasuhito Uehara from \arukikata, Tomoko Ohsuga and Keizo Oyama from National Institute of Informatics in the construction and provision of the Arukikata Travelogue Dataset, the predecessor of the NAIST Academic Travelogue Dataset.
This work was supported by JSPS KAKENHI Grant Number JP22H03648 and JST, PRESTO Grant Number JPMJPR2039.

\section*{Ethical Statements}
The travelogues and travel schedules in this dataset were collected by the first author during the operation of the travelogue posting service, and the copyright of each post is retained by its respective travelogue writer. This dataset is provided to applicants who agree to the terms of use, in accordance with Japanese copyright law.

\bibliography{tacl2021.bib}

@article{osuga2021,
   author	 = "Ohsuga, Tomoko and Oyama, Keizo",
   title	 = {{Sharing Datasets for Informatics Research through Informatics Research Data Repository (IDR) (in Japanese)}},
   journal	 = "IPSJ Transactions on digital practices",
   year 	 = "2021",
   volume	 = "2",
   number	 = "2",
   pages	 = "47--56",
   month	 = "apr"
}

@inproceedings{hao2010,
author = {Hao, Qiang and Cai, Rui and Wang, Changhu and Xiao, Rong and Yang, Jiang-Ming and Pang, Yanwei and Zhang, Lei},
title = {{Equip Tourists with Knowledge Mined from Travelogues}},
year = {2010},
publisher = {Association for Computing Machinery},
booktitle = {Proceedings of the 19th International Conference on World Wide Web},
pages = {401--410},
numpages = {10},
}

@inproceedings{hao2009,
author = {Hao, Qiang and Cai, Rui and Wang, Xin-Jing and Yang, Jiang-Ming and Pang, Yanwei and Zhang, Lei},
title = {{Generating Location Overviews with Images and Tags by Mining User-Generated Travelogues}},
year = {2009},
publisher = {Association for Computing Machinery},
booktitle = {Proceedings of the 17th ACM International Conference on Multimedia},
pages = {801--804},
numpages = {4},
}

@article{pang2011,
        Author = {Yanwei Pang and Qiang Hao and Yuan Yuan and Tanji Hu and Rui Cai and Lei Zhang},
        Journal = {Computer Vision and Image Understanding},
        Number = {3},
        Pages = {352--363},
        Title = {{Summarizing Tourist Destinations by Mining User-Generated Travelogues and Photos}},
        Volume = {115},
        Year = {2011},
}

@article{akehurst2009,
  title={User generated content: the use of blogs for tourism organisations and tourism consumers},
  author={Akehurst, Gary},
  journal={Service Business},
  volume={3},
  number={1},
  pages={51--61},
  year={2009},
  publisher={Springer}
}

@article{hashimoto2011knbc,
	author	=	"Hashimoto, Chikara  and Kurohashi, Sadao and Kwahara, Daisuku and Shinzato, Keiji and Nagata, Masaaki",
	title =		"{{Construction of a Blog Corpus with Syntactic, Anaphoric, and Sentiment Annotations (in Japanese)}}",
	journal = 	"Natural Language Processing",
	pages =		"175--201",
	volume =	18,
	number =	3,
	year = 		2011,
}

@inproceedings{pustejovsky2013capturing,
  title={{Capturing Motion in ISO-SpaceBank}},
  author={Pustejovsky, James and Yocum, Zachary},
  booktitle={Proceedings of the 9th Joint ISO-ACL SIGSEM Workshop on Interoperable Semantic Annotation},
  pages={25--34},
  year={2013}
}

@inproceedings{pustejovsky-etal-2015-semeval,
    title = {{SemEval-2015 Task 8: SpaceEval}},
    author = "Pustejovsky, James  and
      Kordjamshidi, Parisa  and
      Moens, Marie-Francine  and
      Levine, Aaron  and
      Dworman, Seth  and
      Yocum, Zachary",
    booktitle = {{Proceedings of the 9th International Workshop on Semantic Evaluation (SemEval 2015)}},
    year = "2015",
    url = "https://aclanthology.org/S15-2149",
    doi = "10.18653/v1/S15-2149",
    pages = "884--894",
}

@inproceedings{caselli_sprugnoli_dnt2021, 
    title={{DNT: un Corpus Diacronico e Multigenere di Testi in Lingua Inglese}}, 
    author={Caselli, Tommaso and Sprugnoli, Rachele}, 
    booktitle={AIUCD2021 - Book of Abstracts, Quaderni di Umanistica Digitale.}, 
    year={2021}
}
\bibliographystyle{acl_natbib}

\end{document}